\documentclass[letterpaper]{article} % DO NOT CHANGE THIS
\usepackage{aaai25}  % DO NOT CHANGE THIS
\usepackage{times}  % DO NOT CHANGE THIS
\usepackage{helvet}  % DO NOT CHANGE THIS
\usepackage{courier}  % DO NOT CHANGE THIS
\usepackage[hyphens]{url}  % DO NOT CHANGE THIS
\usepackage{graphicx} % DO NOT CHANGE THIS
\usepackage{natbib}  % DO NOT CHANGE THIS AND DO NOT ADD ANY OPTIONS TO IT
\usepackage{caption} % DO NOT CHANGE THIS AND DO NOT ADD ANY OPTIONS TO IT
\usepackage{algorithm}
\usepackage{algorithmic}
\usepackage{newfloat}
\usepackage{listings}
\DeclareCaptionStyle{ruled}{labelfont=normalfont,labelsep=colon,strut=off} % DO NOT CHANGE THIS
\floatstyle{ruled}
\newfloat{listing}{tb}{lst}{}
\floatname{listing}{Listing}
\title{Data Wrangling Task Automation Using Code-Generating Language Models}
\author {
    Ashlesha Akella,
    Krishnasuri Narayanam
}
\affiliations {
    IBM Research India\\
    ashlesha.akella@ibm.com, knaraya3@in.ibm.com
}
\usepackage{bibentry}
\begin{document}

\maketitle

\begin{abstract}
Ensuring data quality in large tabular datasets is a critical challenge, typically addressed through data wrangling tasks. Traditional statistical methods, though efficient, cannot often understand the semantic context and deep learning approaches are resource-intensive, requiring task and dataset-specific training. To overcome these shortcomings, we present an automated system that utilizes large language models to generate executable code for tasks like missing value imputation, error detection, and error correction. Our system aims to identify inherent patterns in the data while leveraging external knowledge, effectively addressing both memory-dependent and memory-independent tasks.
\end{abstract}

\section{Introduction}

Tabular datasets in industrial use cases represent structured data with numerous rows and columns. Since data is pivotal in business decision-making, preserving data quality has become paramount. Data wrangling tasks like missing value imputation, error identification, and error correction are crucial to enhancing the overall data quality of tabular datasets by resolving data inconsistencies and errors.

State-of-the-art data quality enrichment approaches found in the literature face distinct challenges. Traditional statistical methods \cite{van2018flexible,gong2021missing,thomas2021systematic} applied to large tabular datasets efficiently leverage statistical properties of structured data with minimal time and computational requirements, however, they usually fail to account for the semantic context of the data (e.g., impute the column {\em state} given {\em city}). While deep learning approaches \cite{lin2022deep,samad2022missing,huang2024semi} show promise, they require large training datasets and models tailored to specific tasks and datasets, making them time-consuming and resource-intensive to develop and deploy.

Large Language Models (LLMs) offer a novel approach \cite{iida2021tabbie,narayan2022can,huh2023pool,DataWranglingML23,liu2023jarvix,liu2024coachlm,ashlesha2024automatic} to address these challenges in data quality enrichment due to their extensive knowledge gained from training on massive datasets. However, data wrangling with LLMs applied on row-level tabular data can become computationally expensive for large datasets, making it difficult to balance accuracy and efficiency. 

To address this challenge, code-generating LLMs can be leveraged to automatically translate data patterns into concise executable code, enabling efficient tasks like imputation, error detection, and correction. While \cite{li2024towards} work has proposed code generation workflows, they often face limitations that require external knowledge or iterative refinement, highlighting the need for more adaptable systems that incorporate both inherent data patterns and external information.

% Existing code generation workflows \cite{li2024towards} for data-wrangling struggle with tasks requiring external knowledge or iterative code refinement.

Our system uses LLMs to automatically generate code for data-wrangling tasks, incorporating relevant external knowledge as needed. As accommodating numerous columns in a dataset is resource-intensive, the system selects only the columns semantically relevant to the task. This targeted approach enables the LLM to efficiently generate code, allowing the system to execute data-wrangling tasks without additional LLM calls per row, thereby enhancing scalability for large datasets. Moreover, our system employs an iterative refinement technique to optimize the generated code snippets.

\begin{figure}[t]
  \includegraphics[width=0.97\columnwidth]{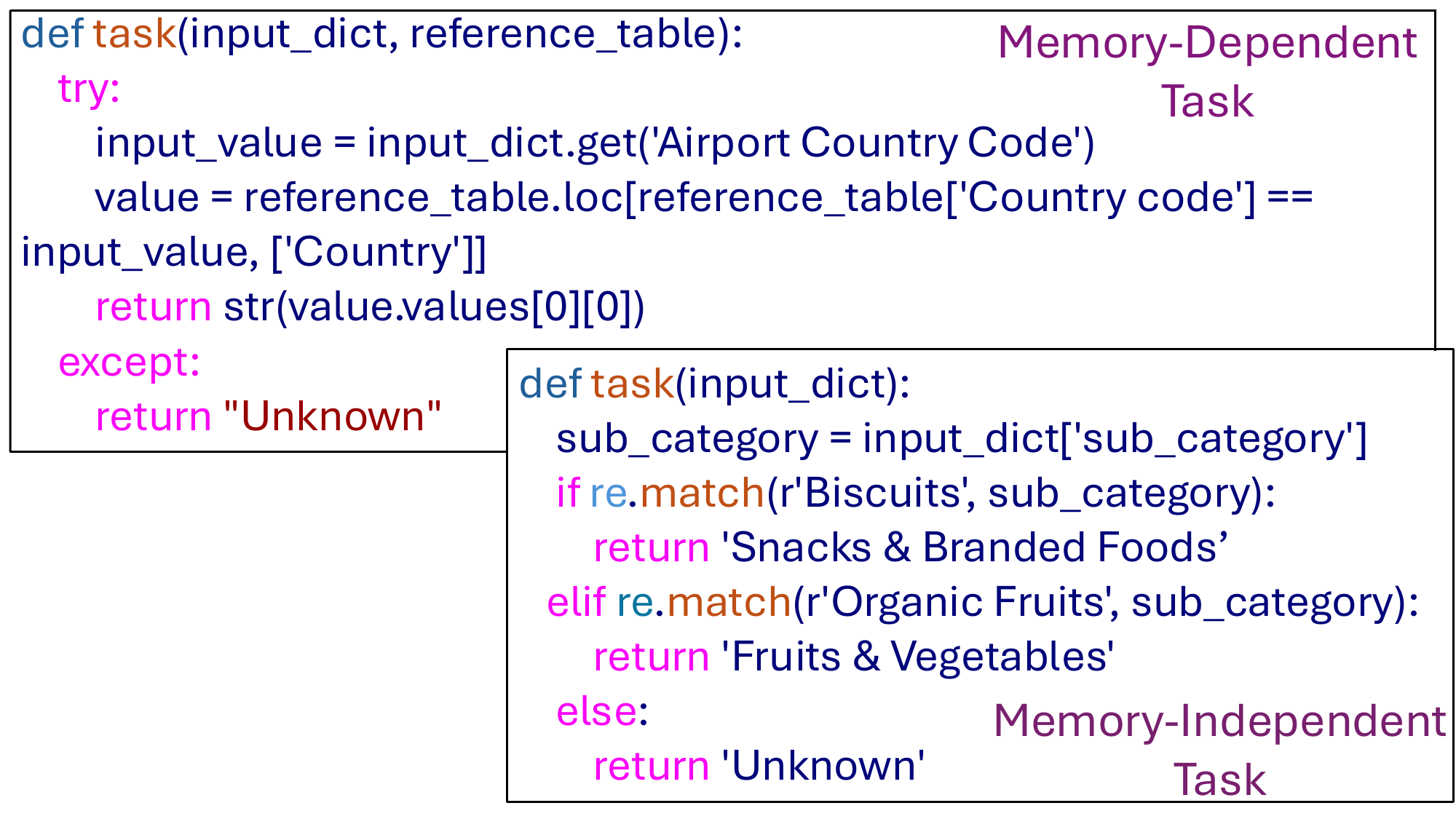}
  \caption{Auto generated code by our system for Missing Value Imputation task. Impute `Category' in {\em BigBasket} dataset (above) and `Country Name' in  {\em Airline} dataset.}
  \label{fig:code_example}
\end{figure}

\section{Leveraging Code-Generating LLMs}

Tabular datasets often contain inherent patterns with dependencies between specific columns. E.g., the `24-Hour Service' column in the `Starbucks' dataset \cite{Starbucks.dataset} can be set to `True' if both `Opening Time' and `Closing Time' are midnight. Such patterns can be represented as concisely formulated executable code. Our system automates code generation for data-wrangling tasks leveraging LLMs trained on extensive code datasets \cite{mishra2024granite}. It reduces manual effort and minimizes repeated LLM calls, enhancing computational efficiency. Figure \ref{fig:code_example} shows an example code generated by our system using an iterative prompting approach. It even allows the LLM to refine its output based on previous responses.

\begin{figure}[t]
  \centering
  \includegraphics[width=1\columnwidth]{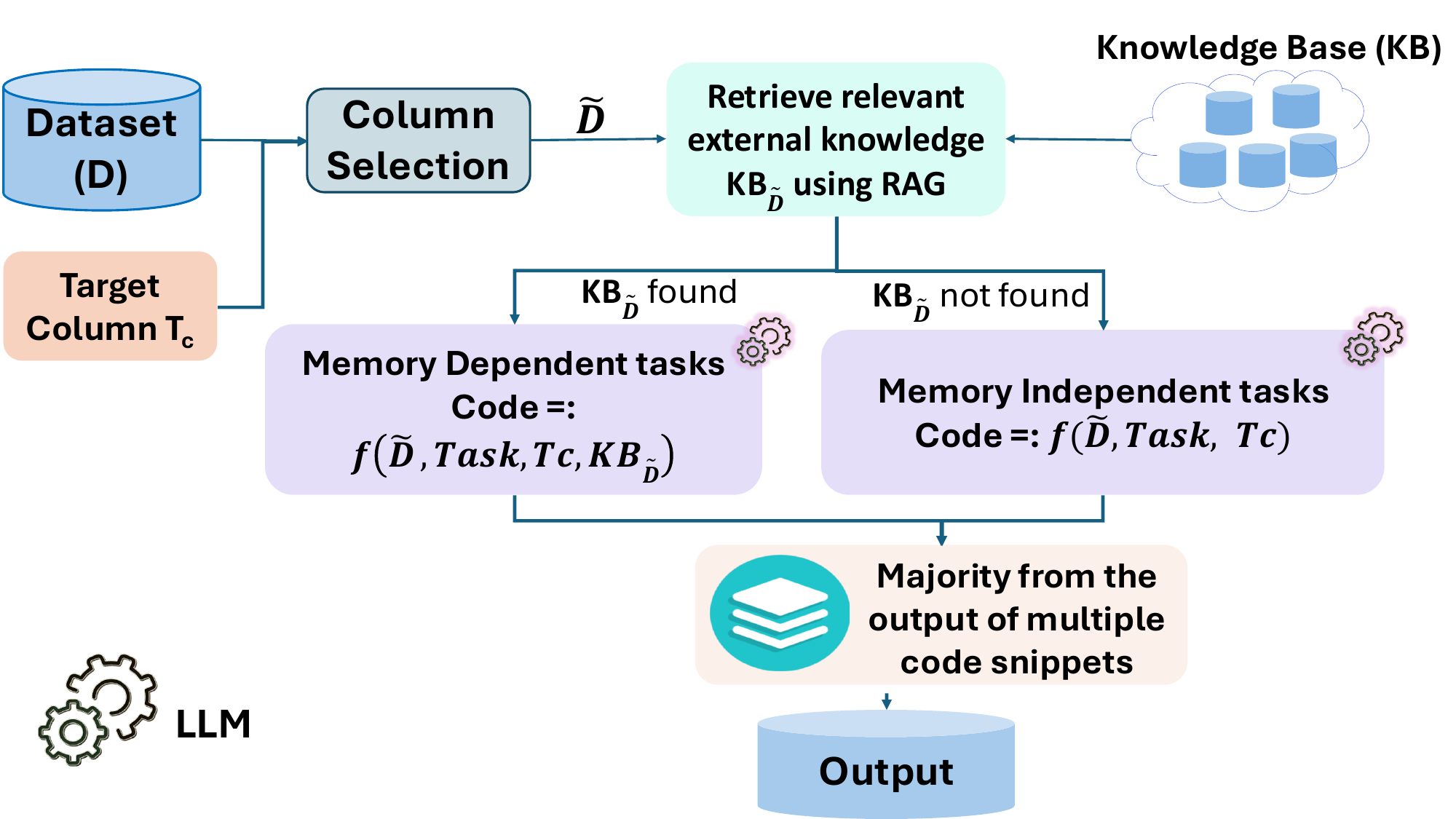}
  %\caption{Workflow of automated code generation for data wrangling tasks}
  \caption{Workflow of data wrangling task automation through code generation}
  \label{fig:system_arch}
\end{figure}

\section{Methodology}

Our system classifies tasks into two primary categories based on the necessary external knowledge:

\begin{enumerate}
    \item \textbf{Memory-Dependent tasks}: They require the incorporation of external or organizational data, such as established mappings (e.g., cities to states) or evolving business mappings (e.g., imputing job role based on job title)
    \item \textbf{Memory-Independent tasks}: They solely depend on the input table and can be further categorized into two groups:
    (i) \textit{Row-alone Tasks}: Data present in the current row is sufficient to accomplish these tasks without any contextual knowledge (e.g., imputing the `24-Hour Service' column using `Opening Time' and `Closing Time' columns). (ii) \textit{Few-shot Tasks}: Such tasks can be inferred using contextual knowledge learnt from a small number of examples provided as part of the prompt.
\end{enumerate}

Figure \ref{fig:system_arch} depicts the overall architecture of our system. Given a task dataset $D$ and a target column, our system employs a Histogram-based Gradient Boosting Classification Tree \cite{guryanov2019histogram} method to identify columns relevant to the target column, resulting in the filtered dataset $\widetilde{D}$. Inspired by the RAG approach \cite{lewis2020retrieval}, our system finds if there is any data in the external Knowledge Base (KB) that is semantically similar to $\widetilde{D}$. If there exists any such data ($KB_{\tilde{D}}$) exceeding a preset similarity threshold, the \textit{Memory-Dependent} workflow is executed using the selected $KB_{\tilde{D}}$; else, the \textit{Memory-Independent} workflow proceeds. Both workflows use iterative prompting for code generation.

\section{Code Generation}
% Depending on the task, users are prompted to identify a target column. For tasks like data imputation and error correction, the system uses ground truth data ($G$) from non-null rows in the target column. For error detection, users supply a small labeled ground truth dataset ($G$) annotated with 'Yes'/'No' labels. The system applies k-fold cross-validation on $G$, using k-1 folds to generate code and the k\textsuperscript{th} fold for validation, allowing iterative code refinement.

%Users begin by selecting a data-wrangling task and a target column.
For data imputation and error correction, the system uses non-null rows in the target column as ground truth $G$. For error detection, users provide a small annotated dataset with `Yes'/`No' labels as ground truth. Our system employs $k$-fold cross-validation on $G$, generating code with $k$-1 folds and validating with the $k^{th}$ fold. It generates multiple code snippets through k-fold cross-validation, each capturing different patterns. Data-wrangling task is executed with each code snippet, and final output is determined through a majority consensus among the outputs of all code snippets.
\textbf{Iterative Prompt Generation}: The prompts for the system are constructed through an iterative process \cite{wang2022iteratively}. The initial prompt contains instructions and a small set of randomly selected rows from the ground truth data $G$. In subsequent iterations, the prompt is enhanced by incorporating the most effective code generated in previous iterations.

In each iteration, for \textit{Memory-Dependent} tasks, code is generated using external data $KB_{\tilde{D}}$ and validated on ground truth $G$. For \textit{Memory-Independent} tasks, system tries the {\em Few-shot} method only after exhausting {\em Row-alone} method.
In {\em Row-alone} method, a few diverse samples are selected from $G$ using K-means clustering algorithm (with samples representing centroids of the clusters) to generate a code snippet, which is applied on $G$. If accuracy of this operation is $\geq 0.9$, then the same code is applied on $D$. In {\em Few-shot} method, one randomly selected sample $r$ from $G$ is provided along with few-shot examples from $G$ that are semantically similar to $r$ to generate a code snippet. We used the LLM model  \textit{granite-34b-code-instruct-8k} \cite{mishra2024granite}.

The prompt is structured with instructions: \textit{Task Description}: Directing the LLM to ``Write a Python code..''. \textit{Function Behavior}: Instructing to detect patterns, and return ``Unknown'' when no pattern is found. \textit{Example Data}: A small randomly selected sample of $n$ rows from the input table, with each row separated by newlines and column values by semicolons. \textit{Example Code}: The most effective code is selected from the previous iteration of k-fold cross-validation. \textit{Reference Table}: For Memory-Dependent tasks, a sample from the reference data $KB_{\tilde{D}}$ is added to the prompt. 

\section{Results}
%The results demonstrate that proposed system achieves comparable performance to row-wise LLM invocation while significantly reducing the number of LLM calls required.
% The results show that our proposed system achieves comparable performance to LLM invocation per row with significant reducing the number of LLM calls.
The results (using Airline \cite{Airline.dataset} 
and BigBasket \cite{BigBasket.dataset} datasets) show that our system achieves comparable performance to a system that calls the LLM for each row, while significantly reducing the number of LLM calls.

\begin{table}[ht]
\centering
\resizebox{.97\columnwidth}{!}{
\begin{tabular}{l|l|l|l}
% \begin{tabular}{p{2.4cm}|p{1.5cm}|p{1.8cm}|p{1.8cm}}
    \hline
    Task & Dataset & row-wise & code-gen\\
    \hline
    Imputation & Airline & 0.97 (\textit{\#1376}) & 0.99 (\textit{\#20})\\
    Error Detection & Airline & 0.98 (\textit{\#1376}) & 0.98 (\textit{\#15})\\
    Error Correction & Airline & 0.98 (\textit{\#1376}) & 0.98 (\textit{\#15})\\
    Imputation & BigBasket & 0.94 (\textit{\#1512}) & 0.94 (\textit{\#30})\\
    Error Detection & BigBasket & 0.94 (\textit{\#1512}) & 0.98 (\textit{\#32})\\
    Error Correction & BigBasket & 0.92 (\textit{\#1512}) & 0.94 (\textit{\#34})
\end{tabular}}
%\caption{Comparison of accuracy between row-wise LLM invocation and code generation (our system), with the total number of LLM calls indicated by \#.}
\caption{Accuracy of row-wise LLM invocation vs. code generation approach (with the number of LLM calls \#)}
\label{table2}
\end{table}

\bibliography{aaai25.bib}

\end{document}